\documentclass{bmvc2k}
\pdfoutput=1

\title{Quantised Transforming Auto-Encoders:
\\ Achieving Equivariance to Arbitrary Transformations in Deep Networks}

\addauthor{Jianbo Jiao}{jianbo@robots.ox.ac.uk}{1}
\addauthor{João F. Henriques}{joao@robots.ox.ac.uk}{1}

\addinstitution{Visual Geometry Group\\
University of Oxford}

\runninghead{J. Jiao, J.F. Henriques}{Quantised Transforming Auto-Encoders}

\usepackage{epsfig}
\usepackage{graphicx}
\usepackage{amsmath}
\usepackage{amssymb}

\def\eg{\emph{e.g}\bmvaOneDot}

\def\etal{\emph{et al}\bmvaOneDot}
\def\ie{\emph{i.e.~}} 
 
\def\etc{\emph{etc.~}}

\usepackage{wrapfig,booktabs}
\usepackage{multirow}
\usepackage{adjustbox}

\renewcommand{\paragraph}[1]{\vspace{2pt plus 1pt minus 1pt}\noindent{\bf #1}\;}

\usepackage{hyperref}
\hypersetup{
        final,
        breaklinks,
        colorlinks,
        linkcolor={red!60!black},
        citecolor={green!60!black},
        urlcolor={blue!60!black}
        }

\newcommand{\R}{\mathbb{R}}      
\newcommand{\Z}{\mathbb{Z}}      
\newcommand{\X}{\mathcal{X}}     
\newcommand{\Y}{\mathcal{Y}}     
\newcommand{\D}{\mathcal{D}}     
\newcommand{\U}{\mathcal{U}}     
\newcommand{\yt}{y_{\mathrm{t}}} 
\newcommand{\ye}{y_{\mathrm{e}}} 
\newcommand{\gray}[1]{\textcolor{gray}{#1}} 

\begin{document}

\maketitle

\begin{abstract}
   In this work we investigate how to achieve equivariance to input transformations in deep networks, purely from data, without being given a model of those transformations. Convolutional Neural Networks (CNNs), for example, are equivariant to image translation, a transformation that can be easily modelled (by shifting the pixels vertically or horizontally). Other transformations, such as out-of-plane rotations, do not admit a simple analytic model. We propose an auto-encoder architecture whose embedding obeys an arbitrary set of equivariance relations simultaneously, such as translation, rotation, colour changes, and many others. This means that it can take an input image, and produce versions transformed by a given amount that were not observed before (\eg a different point of view of the same object, or a colour variation). Despite extending to many (even non-geometric) transformations, our model reduces exactly to a CNN in the special case of translation-equivariance. Equivariances are important for the interpretability and robustness of deep networks, and we demonstrate results of successful re-rendering of transformed versions of input images on several synthetic and real datasets, as well as results on object pose estimation.\footnote{Project page: \url{https://www.robots.ox.ac.uk/~vgg/research/qtae/}.}
\end{abstract}

\section{Introduction}

Deep learning has achieved impressive performance in many pattern recognition tasks, especially in the visual domain~\cite{krizhevsky2012imagenet,goodfellow2016deep}.
A crucial factor in this success is the use of Convolutional Neural Networks (CNNs), which are highly tuned to natural images due to their natural equivariance to translations.
Nevertheless, other transformations are not widely adopted as equivariances in common deep network applications in vision~\cite{lenc2015understanding}, with the closest equivalent being data augmentation~\cite{goodfellow2016deep}, which achieves invariance instead, by randomly transforming input images during training.
When we say an operator is transformation equivariant, it means that the effect of the applied transformation is detectable in the operator output, and its effect has a predictable functional form~\cite{lenc2015understanding}.
When considering the task of imagining what a visual scene looks like from a different point of view or with different components, which is important in computer graphics, planning and counter-factual reasoning, translation equivariance is not sufficient.
There are more transformations needed to manipulate a scene freely, for instance affine transformations, out-of-plane rotation, shape, lighting, \etc
While enabling in CNNs the property of equivariance to various analytical transformations has been explored in the literature (see section~\ref{sec:related}), doing so without an analytical model of the transformation is relatively under-explored.
In this paper, we are interested in the question: \emph{can we make deep networks equivariant to arbitrary transformations from data alone?}

An influential line of work that attempts to address this question (as well as the separate question of compositionality) is capsule networks~\cite{hinton2011transforming}.
They extend auto-encoders with an equivariant latent code, and propose a simple perturbation-based training method to enforce its equivariance w.r.t. example transformations.
A single capsule thus contains an embedding part with the visual entity, and another part encoding the deformations, which can then be manipulated or used for downstream tasks.
Recent works~\cite{hinton2018matrix,kosiorek2019stacked} propose larger capsule networks,
focusing on the composability aspect.
However, one limitation is that they represent the transformation parameters as continuous embeddings, which are limited to simple relationships between image-space and the transformations.

In this paper, we propose to \emph{discretise} the embeddings,
resulting in quantised representations that are orthogonal to each other.
Specifically, we use different \emph{tensor dimensions} to represent different transformations, so each dimension encodes a single transformation parameter independently.
Unlike capsule networks, where transformation parameters are manipulated additively, in our discretised space the embedding is rolled or shifted (depending on whether the parameter is cyclic) by a specific discrete value.
Interestingly, our method \emph{encompasses convolutional neural networks as a special case} -- the case of translation equivariance -- while extending them to more transformations, including non-geometric.
We experimentally show that the proposed method can model various transformations in an equivariant manner, even for out-of-plane 3D transformations without any 3D operations such as 3D convolutions or ray-tracing.
In summary, the main contributions of our work are:

\noindent 1. A method to learn equivariance to arbitrary (non-geometric) transformations from data.

\noindent 2. An exploration of tensor-product and tensor-sum spaces for combining transformations.

\noindent 3. Experimental results across several synthetic and real datasets, demonstrating extrapolation to unseen attribute combinations and pose estimation.

\section{Related Work}\label{sec:related}

There are many notable examples of transformation invariance in computer vision's history, with many consisting of linear models, such as tangent vectors centered on training images~\cite{simard2000transformation}, and on densely-sampled transformed images~\cite{miao_learning_2007,tenenbaum2000separating, henriques2014pose}.
Many works focus on rotation~\cite{schmidt_learning_2012} or scale~\cite{kanazawa2014locally} alone, and most focus on geometric transformations only, as well as linear models.
Building on earlier work on invariance for Boltzmann machines~\cite{memisevic2010learning}, Hinton~\etal~\cite{hinton2011transforming} proposed ``capsules'', which were meant to generalize CNN's filters, recognizing patterns not only across translation, but also across other transformations.
While this work studied capsules in isolation, more recent work focused on composing them together~\cite{hinton2018matrix}.
They were also extended to unsupervised learning, achieving high performance in image classification~\cite{kosiorek2019stacked}.
Lenc and Vedaldi~\cite{lenc2015understanding} studied the emergent invariance and equivariance in networks trained for classification.
Geometric transformations can be seen as forming a group, which allows analysis with tools from abstract algebra, and this has been used successfully in many works~\cite{bruna2013learning,henriques2017warp,kondor2018generalization,CohenNEURIPS2019,qi2020learning,s2018spherical}; usually only geometric or rigid transformations are considered.
In addition to obtaining equivariant representations, it is frequently desired that they are disentangled, which informally means that they can be manipulated independently~\cite{higgins2018towards,kulkarni2015deep,xiao2019identity}.
Image synthesis is also extensively studied in computer vision and graphics.
Some works~\cite{chaurasia2013depth,kopf2013image,ladicky2014pulling,penner2017soft,rematas2016novel,seitz2006comparison} compute 3D representations of the scene explicitly, while others handle the 3D geometry with implicit functions~\cite{fitzgibbon2005image,matusik2002image}.
Recent deep learning-based approaches~\cite{flynn2019deepview,flynn2016deepstereo,hedman2018deep,mildenhall2019local,srinivasan2019pushing,eslami2018neural,sun2018multi} can fill in missing data such as occlusions or holes, though most rely on multi-view inputs and so are more restricted.
Several methods~\cite{dosovitskiy2016learning,park2017transformation,tatarchenko2016multi,yang2015weakly,zhou2016view,kulkarni2015deep,sitzmann2019srns} directly learn a mapping from a source image to the target image, purely from data, which is closer to ours.
However, these methods usually need additional depth sensor as input or supervision, and do not address transformations beyond geometric ones.
A recent approach is to render scenes by ray-tracing~\cite{mildenhall2020nerf}, though generally only viewpoint transformations are considered, as opposed to general latent spaces in auto-regressive~\cite{van2016pixel} or adversarial~\cite{goodfellow2014generative} models.
Equivariance can be a complementary objective to enforce geometric structure during learning, leveraged for example by Dupont \etal~\cite{dupont2020equivariant} to improve neural rendering, or symmetry to reflections used to learn 3D structure~\cite{Wu_2020_CVPR}.

\begin{figure}
    \centering
    \includegraphics[width=0.85\textwidth]{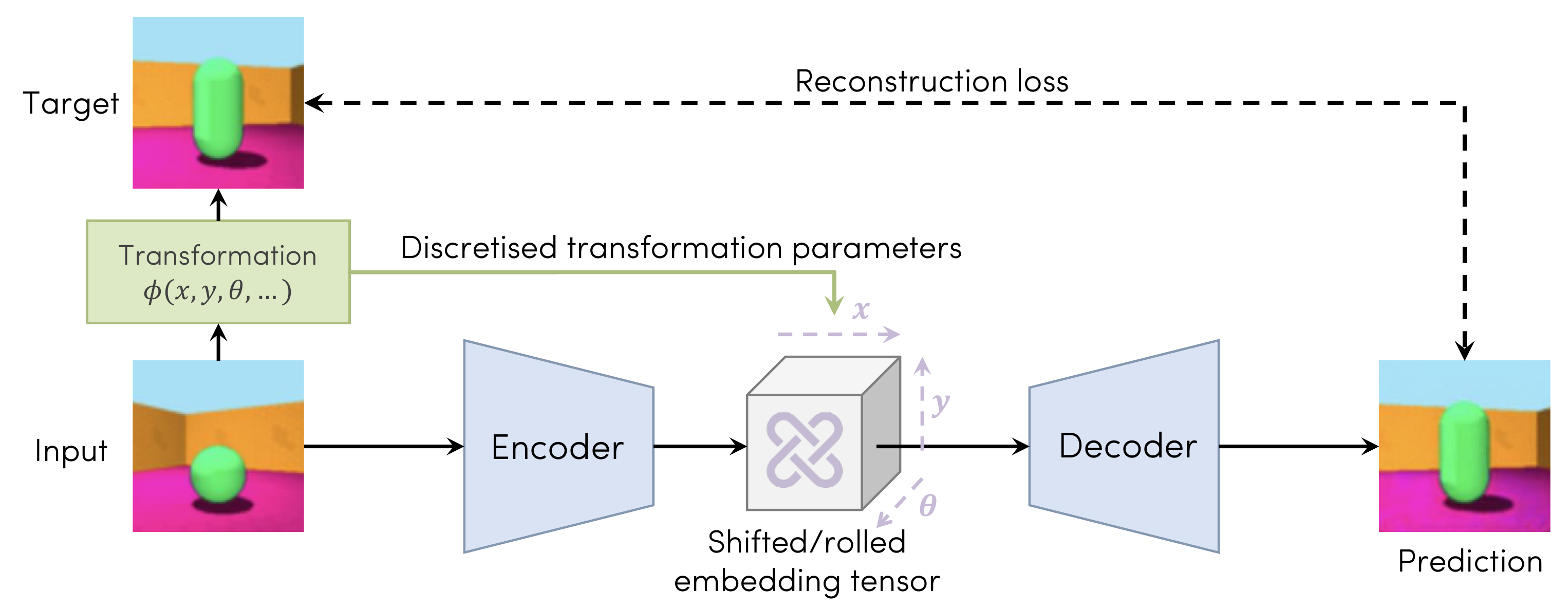}
    \caption{Our method trains a network with pairs of images related by a transformation (which may include arbitrary non-geometric changes). An embedding tensor that encodes the image is shifted along different dimensions according to the transformed amounts, and the decoded image must match the corresponding transformed image (target). Refer to section~\ref{s:qae} for a full description.}
\label{fig:frame}
\end{figure}

\section{Method}

\subsection{Background}

Consider training a classical auto-encoder, where the goal is to learn the two functions (or their parameters), an encoder $\phi:\,\X\mapsto\Y$ and a decoder $\psi:\,\Y\mapsto\X$, for an input-space $\X$ (\eg images) and an embedding-space $\Y$ (\eg $\R^{m}$). This is learned from a dataset $\D$, so that encoding and decoding a sample results in approximately the same sample:

\begin{equation}
\min_{\phi,\psi}\mathbb{E}_{x\sim\D}\left\Vert \psi(\phi(x))-x\right\Vert .\label{eq:ae}
\end{equation}

The bottleneck of the embedding space encourages the auto-encoder to compress the input distribution.
Many methods improve this bottleneck with regularisation terms or noise~\cite{kingma_auto-encoding_2014}.

\subsection{Transforming auto-encoders}\label{sec:transae}

A transforming auto-encoder, also known as a capsule network~\cite{hinton2011transforming}, is a modification of eq. \ref{eq:ae} that additionally forces part of the embedding-space to encode a transformation $T:\,\X\times\R^{t}\mapsto\X$ with $t$ parameters. For example, $T$ may perform planar rotation of the image by a given angle ($t=1$), or an affine transformation with $t=6$ parameters. The auto-encoder's embedding $y\in\Y$ can then be written as the concatenation $y=(\yt,\ye)$, where $\yt\in\R^{t}$ expresses transformation parameters that are specific to the image (\eg how much it is rotated away from a canonical view), and $\ye\in\R^{m-t}$ encodes the rest of the image information (\eg appearance that is not related to the transformation).

The equivariance relation that we would like to (approximately) enforce is:
\begin{equation}
\phi(T(x,u))=\phi(x)+u.\label{eq:equiv}
\end{equation}

Eq. \ref{eq:equiv} establishes an equivalence between translating by $u$ in the embedding-space, and using $T$ to transform an input $x$ by the same amount $u$. To avoid cumbersome math, $u\in\R^{t}$ is only added to the first $t$ elements of the embedding (\ie to $\yt$, not to the full embedding $y$), in a slight abuse of notation.

The transforming auto-encoder is trained by minimising the reconstruction error between both sides of eq. \ref{eq:equiv}, in image-space, for values of $u\in\R^{t}$ randomly sampled from a distribution $\U$ (for example, a uniform range of rotation angles):
\begin{equation}
\min_{\phi,\psi}\;\mathbb{E}_{x\sim\D,u\sim\U}\left\Vert \psi(\phi(T(x,u))-u)-x\right\Vert .\label{eq:tae}
\end{equation}

In practice, $x$ and $T(x,u)$ can be two views of the same scene, with a relative transformation of $u$ (for example, the relative pose between the two views in 3D space).

Eq. \ref{eq:tae} can be used to learn simple equivariances, however achieving the desired linearity with respect to $u$ is difficult in practice. Implicit in this objective is a regression task: setting part of the embedding to a precise value that is equal to the transformation $u$, up to an additive constant. Since deep networks usually excel at classification tasks but are more difficult to train for regression tasks, it makes sense to replace this regression task (continuous representation of the transformation) with classification (discrete or quantised representation).

\subsection{Quantised transforming auto-encoders}\label{s:qae}

We do this by redefining the auto-encoder's embedding as a tensor with one additional dimension per transformation parameter: $y\in\R^{d_{1}\times\cdots\times d_{t}\times m}$, for a total of $t+1$ dimensions. They represent a quantised version of $\R^{t}$, or discrete lattice, with resolution $d_{i}$ for the $i$th dimension. Note that for the special case $y\in\R^{d_{1}\times d_{2}\times m}$ this reduces to a standard CNN embedding, with two spatial dimensions and $m$ channels. Similarly to eq. \ref{eq:equiv}, we define an equivariance relation, but with translation in $\R^{t}$ replaced with translation by $u$ in the discrete lattice:
\begin{equation}
\phi(T(x,u))=S(\phi(x),u),\quad S_{v}(y,u)=y_{u+v}, \forall u,v\in\Z^{t}, \label{eq:equiv-shift}
\end{equation}
where $S(y,u)$ translates or \emph{shifts} vector $y$ by an amount $u$.
For example, if $y=[y_1, y_2, \ldots]$, then $S(y,1) = [y_2, y_3, \ldots]$.
When applied to standard CNNs, eq. \ref{eq:equiv-shift} describes their natural equivariance: translating the input by an amount $u$ results in an equal translation of the activations.\footnote{A stride (subsampling) of $k$ can be taken into account by considering that $S$ translates by $ku$.} Eq. \ref{eq:equiv} does not apply to CNNs, since it only describes continuous parameter-spaces. However, our description extends this concept to more general transformations, including their combination with translation. Replacing this in eq. \ref{eq:tae}:
\begin{equation}
\min_{\phi,\psi}\;\mathbb{E}_{x\sim\D,u\sim\U}\left\Vert \psi(S(\phi(T(x,u)),-u))-x\right\Vert, \label{eq:objective}
\end{equation}
which is superficially similar to eq. \ref{eq:tae}, but the intermediate operations deal with a larger embedding which has more structure (the discrete lattice).
A pipeline of the proposed method when implemented as an encoding-decoding process is shown in figure~\ref{fig:frame}.

\paragraph{Transformations with periodic and aperiodic domains.}
The abstract description of the shift operator (eq. \ref{eq:tae}) ignores how boundary conditions are treated. Consider a transformation with a single parameter, a rotation angle $u\in[0,2\pi]$. To take into account the periodicity of the angular domain, the shift operator would be defined as
$S_{v}(y,u)=y_{(u+v)\textrm{ mod }2\pi},$
where mod is the modulus operation. For ease of notation, we assume that the tensor (in this example, vector) $y$ can be indexed with non-integer values, mapping the values in $[0,2\pi]$ to its $d$ elements in ascending order. This reasoning applies to all rotational parameters and the operation is more accurately described as \emph{rolling} instead of shifting. Similarly, for transformation parameters that are not periodic, the shift with boundary conditions would be
\begin{equation}
S_{v}(y,u)=\begin{cases}
y_{u+v} & \textrm{if }0\leq u+v<d\\
0 & \textrm{otherwise}.
\end{cases}\label{eq:shift-boundary}
\end{equation}

In general, one may combine any number of transformation parameters, periodic or not, and with different discretisation size $d$, into a composite shift $S$ that shifts or rolls each dimension independently.
In standard CNNs, eq.~\ref{eq:shift-boundary} corresponds to horizontal and vertical shifts, which can be applied independently, and are zero-padded when out-of-bounds.

\paragraph{Combining multiple transformations.}
The formulation from eq. \ref{eq:shift-boundary} suggests a direct way to handle multiple transformations: apply $t$ shifts to different dimensions of a tensor $y\in\R^{d_{1}\times\ldots\times d_{t}\times m}$, where $m$ contains channels that are not transformed. Formally,
\begin{equation}
S_{v_{1},\ldots,v_{t}}^{\otimes}(y,u)=\begin{cases}
y_{u_{1}+v_{1},\ldots,u_{t}+v_{t}} & \textrm{if }0\leq u_{i}+v_{i}<d_{i}\quad\forall i\\
0 & \textrm{otherwise},
\end{cases}\label{eq:multi-transf}
\end{equation}
which is straightforward to adapt for a mix of periodic and aperiodic domains. One drawback is that the memory use grows exponentially with each transformation, which we address next.

\paragraph{Efficient combinations of transformations.}
Rather than having a tensor product of transformations, we consider an additive product: the activations tensor is instead $y\in\R^{(d_{1}+\ldots+d_{t})\times m}$, and the shift operator becomes
\begin{equation}
S^{\oplus}(y,u)=\left[S(y_{i},u_{i})\right]_{i=1}^{t},
\label{eq:effi}
\end{equation}
where $\left[\cdot\right]_{i=1}^{t}$ stacks $t$ matrices vertically, and $y_{i}$ is the $i$th block of $y$, corresponding to the $i$th transformation (i.e. $y=\left[y_{i}\right]_{i=1}^{t}$). Thus memory scales linearly with the number of transformations, but the activations $y$ can no longer model interactions between transformations.

\paragraph{Interpretability and theoretical justification.}
Although the feature spaces of deep networks are notoriously difficult to interpret, it is natural to ask whether shift-equivariant representations (eq.~\ref{eq:equiv-shift} and eq.~\ref{eq:multi-transf}) take on a recognizable shape.
Interestingly, there is one case that is well-studied: rotation and scaling, which is equivariant to shifts under a log-polar warping of the image~\cite{reddy1996fft}. Concretely, eq.~\ref{eq:equiv-shift} is true for scale and rotation $T$ if $\phi$ is a log-polar warp of the input image $x$.
It has been shown~\cite{henriques2017warp} that the same result holds if $\phi$ is composed with a CNN, and that equivariance extends to different geometric transformations by considering different warps.
The significance of our proposal is that non-linear operations (in the form of a deep network) can achieve arbitrary transformations (including colour and out-of-plane rotations), instead of being limited to geometric transformations of an image.
One limitation of using additive or product tensor-spaces is that they are commutative (\ie the order of transformations does not matter), so in theory they should only model commutative transformations.
This property is shared with the original transforming auto-encoder~\cite{hinton2011transforming}, and yet both methods cope well with affine transformations, which are not commutative (section~\ref{sec:rerendering}).
We speculate this is due to the $m$ non-transformation channels, which can convey arbitrary image content, and can be leveraged by the networks to overcome this difficulty.

\section{Experiments}
Given that the aim of our proposal is to achieve equivariance to generic transformations, and we train an auto-encoder, we focus on \emph{image re-rendering}: the ability of the model to reproduce an input image but with changed transformation parameters.
We test our method's ability to extrapolate beyond the training set for several transformations: affine (including translation and rotation), colour (foreground and background), out-of-plane rotations, object identity changes, and translations in 3D space.
We also assess the method's ability to discriminate poses in these transformation spaces.

\paragraph{Implementation details.}
We use a standard auto-encoder architecture consisting of 4 convolutional and 4 deconvolutional layers (details in supplementary material).
The encoder's embeddings are reshaped to obtain the equivariant tensor ($d_{1}\times\ldots\times d_{t}\times m$), which is then shifted or rolled (eq.~\ref{eq:multi-transf}), and reshaped back to the original dimensions.
An overview is in figure~\ref{fig:frame}.
Our method optimizes eq.~\ref{eq:objective} (with a simple L1 metric), using a product space of transformations (eq.~\ref{eq:multi-transf}) while the Transforming Auto-encoder~\cite{hinton2011transforming} baseline uses eq.~\ref{eq:tae}.
All networks are trained with Adam for 100 epochs, choosing the best learning rate from $\{10^{-5}, 10^{-4}, 10^{-3}\}$.
For a self-contained description of all implementation details please refer to the supplementary material.

\paragraph{Metrics.}
To quantitatively evaluate the performance, we report results using two metrics: 1) Peak Signal to Noise Ratio (PSNR), which describes the accuracy in pixel-space; 2) Structural SIMilarity (SSIM), which more strongly correlates with perceptual similarity.

\paragraph{Baselines.}
We compare with several state-of-the-art view synthesis methods.
The Transforming Auto-Encoder (\emph{Trans.AE})~\cite{hinton2011transforming} is the closest work to ours (see section~\ref{sec:transae} for details).
\emph{MV3D}~\cite{tatarchenko2016multi} proposes to encode both the input image (source view) and the transformation (rotation angle in their case) into latent codes and directly concatenate them, followed by a decoder to predict the target view together with a depth map.
\emph{TVSN}~\cite{park2017transformation} instead generates a flow from the concatenated features of source view and the rotation angle, followed by a hallucination and refinement step.
Similarly, \emph{Chen~\etal}~\cite{chen2019monocular} also generate intermediate outputs of depth and flow, via which the source view is bilinearly sampled to get the target view. Instead of concatenation, the latent code is directly multiplied with the ground-truth viewpoint transformation.
Equivariant Neural Rendering (ENR)~\cite{dupont2020equivariant} is an auto-encoder that similarly obtains an equivariant latent-space, but focuses on viewpoint equivariance; thus it ignores other transformations in our experiments.
To provide a contrast with standard generative models, we also include the $\beta$-VAE~\cite{higgins2016beta} (which generalises VAEs). Note that this model is not trained to achieve equivariance, so it only reconstructs the input image after projecting it to the latent space.
For a fair comparison, all methods are adapted to use the same backbone architecture as ours, and use the same optimiser and training budget.

\begin{table*}
    \centering
    \caption{Quantitative evaluation (PSNR$\uparrow$ and SSIM$\uparrow$) with comparison to the baseline model on four datasets. \textcolor{gray}{Gray} colour means that the method cannot be directly applied to this scenario, and was adapted to incorporate these transformations. $\dagger$As the $\beta$-VAE model is not equivariant, we here only show its self-reconstruction performance without transformations. * indicates that the 3 colour transformations (\emph{wall, floor} and \emph{object} colours) are not changed.}
    \label{tab:quant}
    \resizebox{\columnwidth}{!}{
    \begin{tabular}{@{}lcccccccccccc@{}}
    \toprule
    \multirow{2}{*}{Method} &
      \multicolumn{2}{c}{MNIST~\cite{lecun1998gradient}} &
      \multicolumn{2}{c}{3D Shapes~\cite{3dshapes18}*} &
      \multicolumn{2}{c}{3D Shapes~\cite{3dshapes18}} &
      \multicolumn{2}{c}{SmallNORB~\cite{lecun2004learning}} &
      \multicolumn{2}{c}{RGBD-Object~\cite{lai2011large}} &
      \multicolumn{2}{c}{KITTI~\cite{Geiger2012CVPR}}  \\
     &
      \multicolumn{1}{c}{PSNR} &
      \multicolumn{1}{c}{SSIM} &
      \multicolumn{1}{c}{PSNR} &
      \multicolumn{1}{c}{SSIM} &
      \multicolumn{1}{c}{PSNR} &
      \multicolumn{1}{c}{SSIM} &
      \multicolumn{1}{c}{PSNR} &
      \multicolumn{1}{c}{SSIM} &
      \multicolumn{1}{c}{PSNR} &
      \multicolumn{1}{c}{SSIM} &
      \multicolumn{1}{c}{PSNR} &
      \multicolumn{1}{c}{SSIM} \\ \midrule
    Trans. AE~\cite{hinton2011transforming} & 19.85 & 0.8413 & 28.68 & 0.8447 & 7.77 & 0.3142 & 21.21 & 0.6447 & 19.46 & 0.5520  & 10.23 & 0.3437 \\
    MV3D~\cite{tatarchenko2016multi} & \gray{20.09} & \gray{0.8640} & \gray{26.83} & \gray{0.8812} & \gray{13.33} & \gray{0.6406} & \gray{22.79} & \gray{0.7813} & 19.38 & 0.5612 & 8.85 & 0.3311 \\
    TVSN~\cite{park2017transformation} & \gray{14.66} & \gray{0.3046} & \gray{28.53} & \gray{0.8762} & \gray{20.38} & \gray{0.8283} & \gray{19.11} & \gray{0.7094} & 19.73 & 0.5715 & 7.65 & 0.3141 \\
    Chen \etal~\cite{chen2019monocular} & \gray{20.31} & \gray{\textbf{0.8731}} & \gray{25.10} & \gray{0.8616} & \gray{13.56} & \gray{0.6334} & \gray{17.01} & \gray{0.6768} & 19.56 & 0.5406 & 16.76 & 0.5297 \\
    ENR~\cite{dupont2020equivariant} & - & - & \gray{14.05} & \gray{0.5690} & \gray{7.83} & \gray{0.3884} & \gray{16.67} & \gray{0.6646} & 17.11 & 0.4708 & \gray{6.68} & \gray{0.2357} \\
    $\beta$-VAE~\cite{higgins2016beta}$\dagger$ & \gray{10.22} & \gray{0.3437} & \gray{14.19} & \gray{0.6042} & \gray{8.50} & \gray{0.4081} & \gray{16.22} & \gray{0.6764} & \gray{16.45} & \gray{0.4411} & \gray{12.45} & \gray{0.2662} \\
    Ours     & \textbf{22.40} & 0.8639 & \textbf{30.75} & \textbf{0.9210} & \textbf{28.00} & \textbf{0.8845} & \textbf{24.99} & \textbf{0.8148} & \textbf{19.74} & \textbf{0.5769} & \textbf{16.88} & \textbf{0.5304} \\ \bottomrule
    \end{tabular}
    }
\end{table*}

\subsection{Re-rendering under novel views and transformations}\label{sec:rerendering}

\begin{figure}
    \centering
    \begin{minipage}{.49\textwidth}
        \centering
        \includegraphics[width=\linewidth]{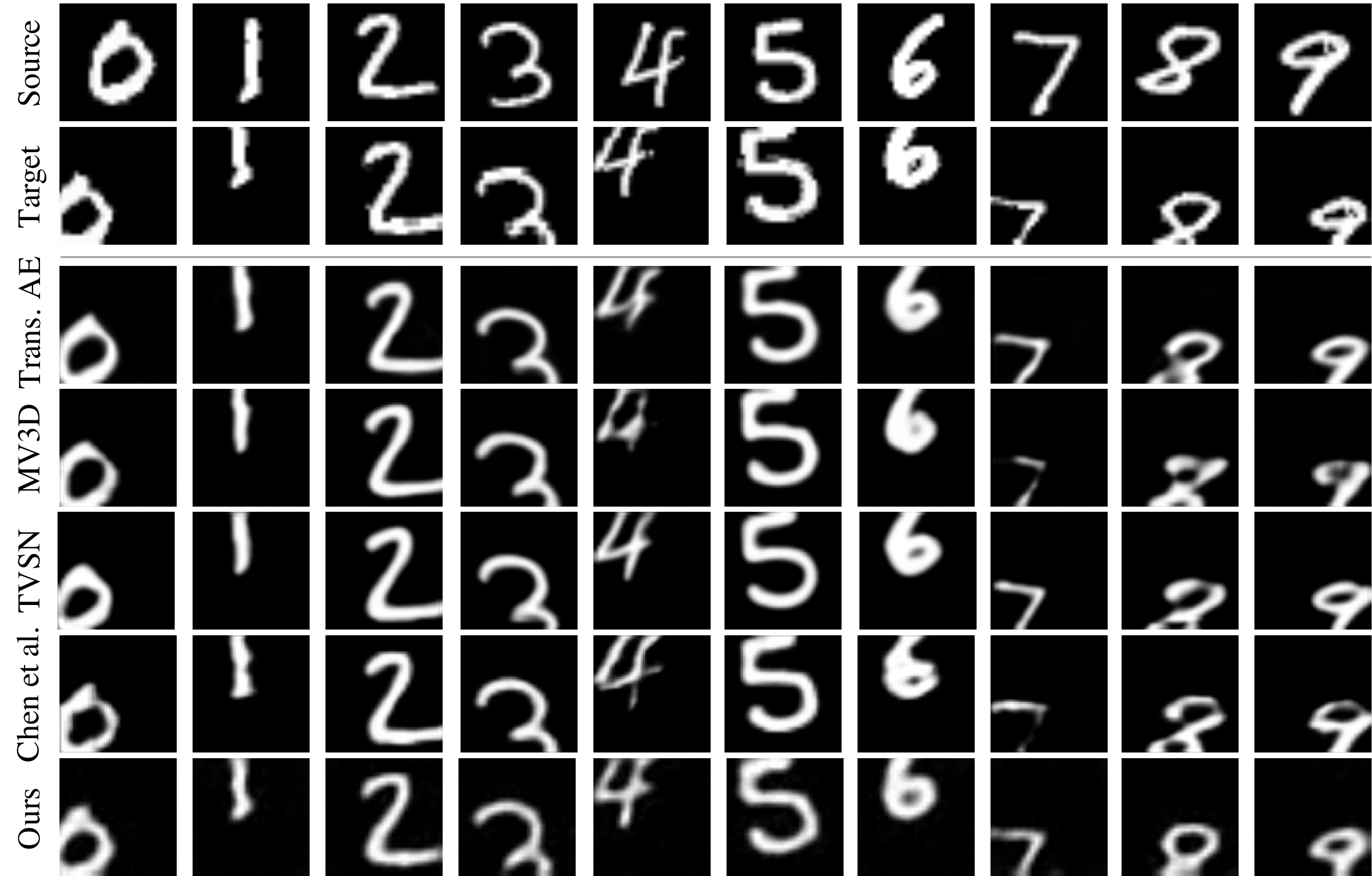}
        \caption{Qualitative performance on the MNIST dataset, with comparison to state-of-the-art methods. Different examples are shown in each column.}
        \label{fig:mnist}
    \end{minipage}
    \hfill
    \begin{minipage}{.49\textwidth}
        \centering
        \includegraphics[width=\linewidth]{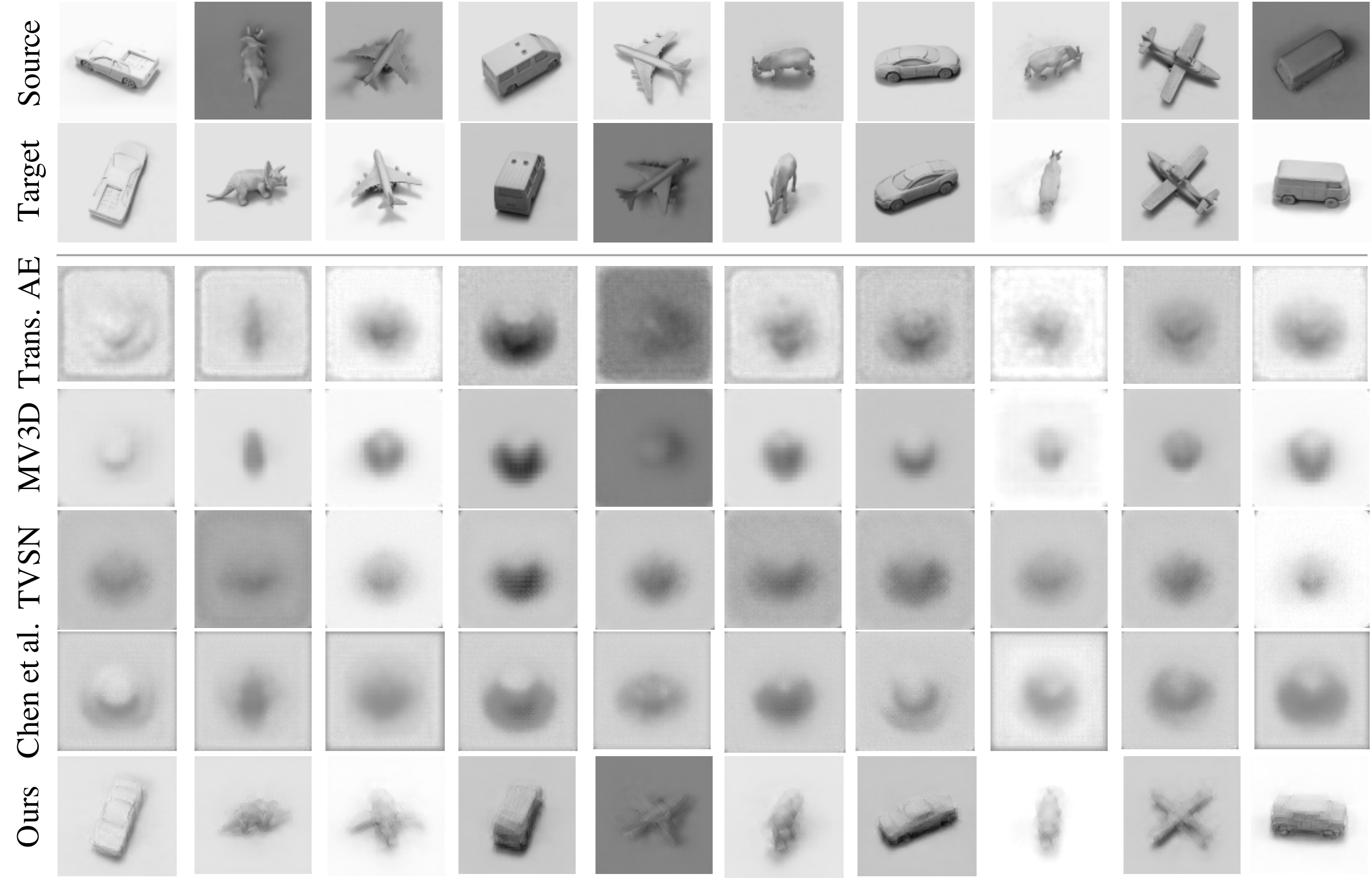}
        \caption{Qualitative results on SmallNORB, compared to state-of-the-art methods. Different combinations of transformations are shown in each column.}
        \label{fig:norb}
    \end{minipage}
\end{figure}

\paragraph{Affine-transformed MNIST.}\label{sec:mnist}
We adapted MNIST~\cite{lecun1998gradient} by applying a random affine transformation to each of the 70K images, maintaining the original training-validation split and image size.
The transformation ranges are $\pm 21$ for translation, $\pm 15^\circ$ for rotation, $[1,1/9]$ for scale, and $\pm 11$ for shear.
We show the results of our method, as well as of all baselines, in figure~\ref{fig:mnist}.
We observe that all methods are able to model the affine transformations and place the digits in the correct place, although with varying loss of quality and detail.
The visualized digits ``4'' and ``7'', in particular, have thin structures that are lost by some methods, and only our method avoids nearly transforming the ``8'' into a ``3''.
Quantitative performance is presented in table~\ref{tab:quant} (second column), confirming that our method and Chen \etal's~\cite{chen2019monocular} perform best overall.

\begin{figure}\centering
    \begin{minipage}{.62\textwidth}
    \centering
    \includegraphics[width=\columnwidth]{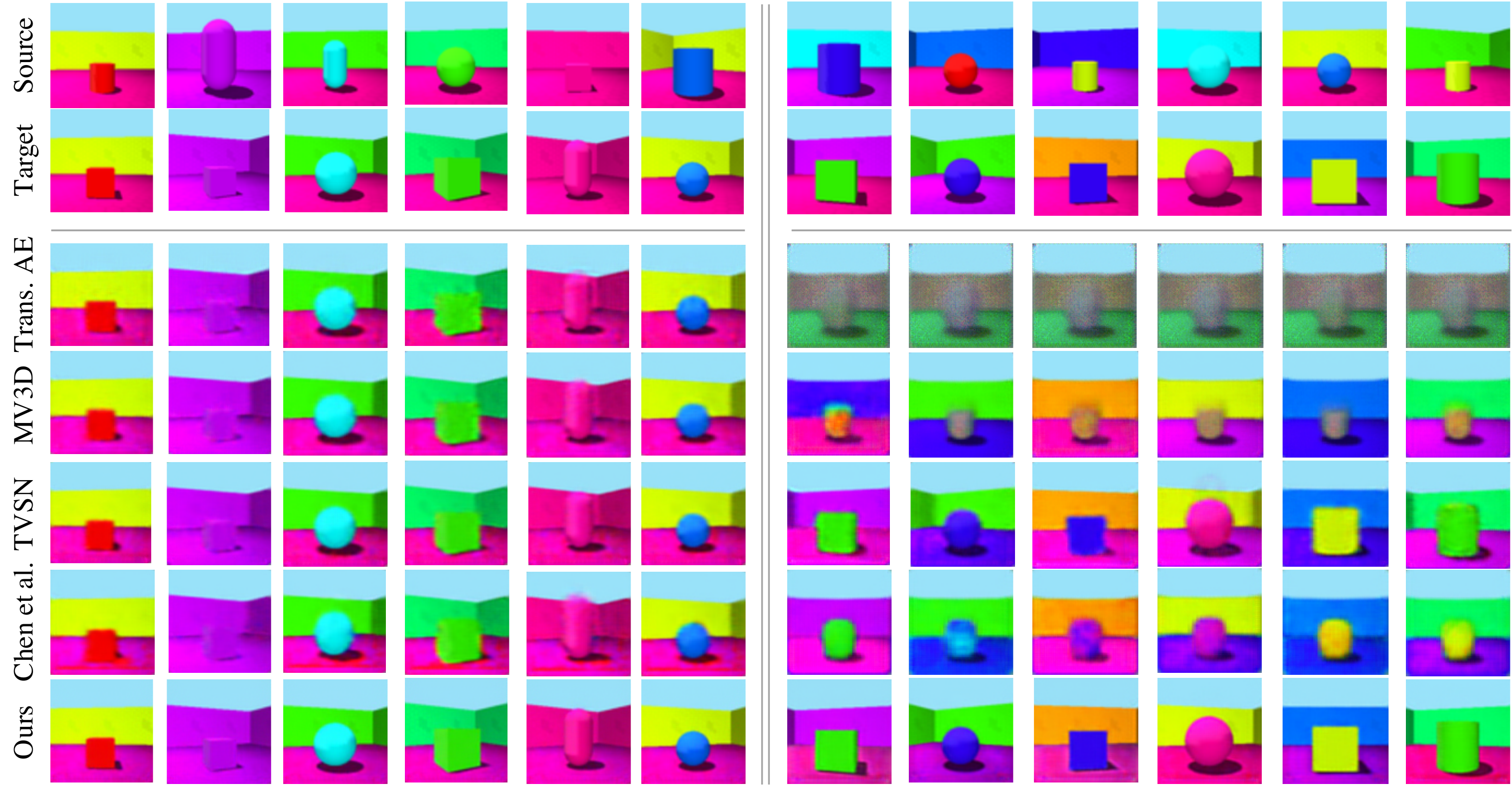}
    \caption{Qualitative performance on the DeepMind 3D Shapes dataset, with comparison to state-of-the-art methods. Different examples are shown in each column. The left part shows the results where the 3 colour related transformations are not used, while the right part are those including the colour transformations.}
    \label{fig:3dshape}
    \end{minipage}
    \hfill
    \begin{minipage}{.37\textwidth}
    \centering
   \includegraphics[width=\columnwidth]{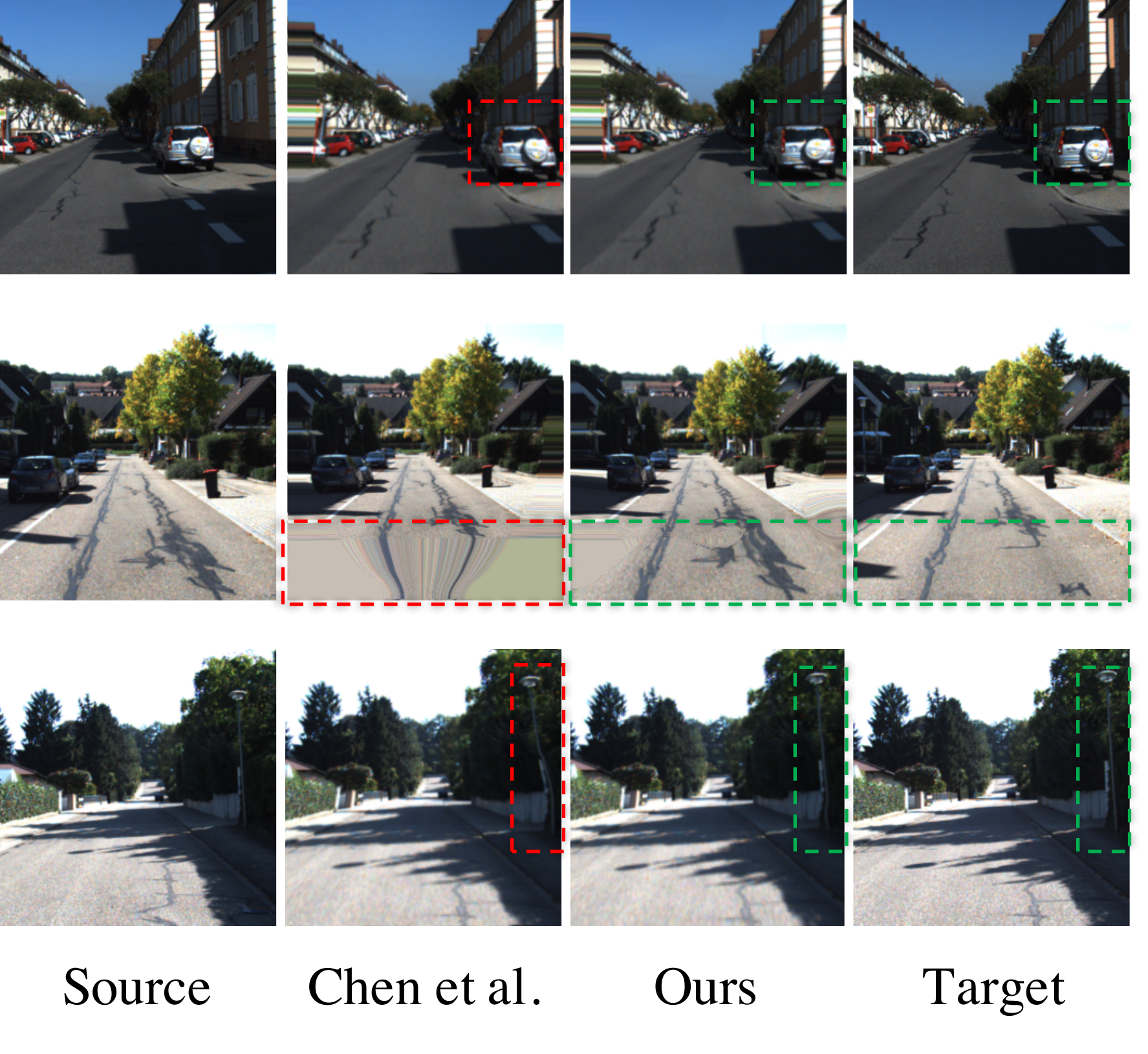}
   \caption{Qualitative performance on the KITTI dataset. Some methods do not generalize to this dataset and are omitted. Regions of interest are highlighted.}
\label{fig:kitti}
\end{minipage}
\end{figure}

\paragraph{DeepMind 3D Shapes.}\label{sec:3dshape}
We now move to a more challenging dataset, DeepMind 3D Shapes \cite{3dshapes18}, which introduces non-geometric transformations (colour changes) and non-planar geometric changes (3D rotation, shape changes).
It consists of 480,000 synthetic RGB images of size $64\times64$, generated by varying 6 factors: \emph{floor colour, wall colour, object colour, scale, shape and orientation}; with 10, 10, 10, 8, 4, and 15 possible values for each factor, respectively.
We consider all factors as transformations in our experiments, and training is as before.
The results are shown in figure~\ref{fig:3dshape} (right panel).
Our method is able to very accurately re-synthesize the image with the given color, shape and rotation changes, while the other methods are much less successful.
For example, a standard transformation auto-encoder (Trans. AE), without our proposed discretisation, falls back to predicting a mean image in all cases.
While forcing the embedding to vary linearly with each transformation is enough for simple affine-warped MNIST images (as in the original capsules proposal~\cite{hinton2011transforming}), our proposed solution of shifting the tensors instead can support much more complex image transformations.
Since we realized that this difficulty may be due to the abrupt changes in colours in the dataset, we ran the same experiment again but with all colours fixed (\ie the transformations are purely geometric, with no colour changes).
The performance of the baseline methods is largely recovered in this case, as shown in figure~\ref{fig:3dshape} (left panel).
This illustrates the fact that many state-of-the-art methods are geared towards geometric transformations, and do not model more general appearance changes correctly.
Quantitative results are in table~\ref{tab:quant}, and additional video results in can be found in the supplementary material.

\begin{figure*}
    \includegraphics[width=\textwidth]{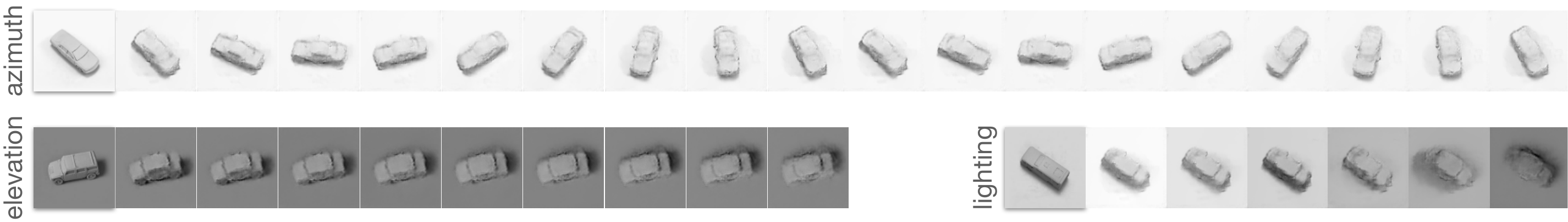}
    \caption{Experiment on SmallNORB for novel view synthesis with isolated transformations. For each transformation, the first image is the input source image, while the images on the right side are the predicted novel views.}
\label{fig:norb_vid}
\end{figure*}

\paragraph{SmallNORB.}
Proceeding to real images, although in a controlled environment, we also tested SmallNORB~\cite{lecun2004learning}.
It contains grayscale images from 50 toys in 5 categories, under 6 lighting conditions, and 3D rotations (9 elevations and 18 azimuths).
We therefore take these three factors as the transformation space.
We use the official train/test split (5 instances of each category for training, and 5 for testing, totalling 24,300 images per set).
From the results (figure~\ref{fig:norb}), we see that SmallNORB is much harder to reconstruct with retargetted poses and lighting for most methods, with all but ours returning an average image, and only ours, Trans.AE and MV3D transforming the lighting correctly (which is non-geometric).
The quantitative results (table~\ref{tab:quant}) reflect this result.
To test our method's ability to differentiate very fine-grained transformations, we plot results varying a single transformation parameter at a time in figure~\ref{fig:norb_vid}.
Note that these are novel views -- only the first image is given, and it is not part of the training set.
Videos with more transitions are in the supplementary material.

\paragraph{RGBD-Object.}
To test re-rendering of unconstrained real-world objects, we used RGBD-Object~\cite{lai2011large}.
It consists of 300 objects captured with a PrimeSense RGBD camera, though we did not use the depth channel.
The objects are organised into 51 semantic categories and in total have 45,000 images, each associated with a single view angle, which we use as the only transformation parameter.
Since there are few instances per category, we test with one instance per category and use the remainder for training.
We show quantitative results in table~\ref{tab:quant}, which show that the images reconstructed by our method are closer to the ground truth.
Due to lack of space the qualitative results are shown in the supplementary material, but we observe that reconstructions for all methods are less detailed than in previous datasets, which is explained by the higher complexity of the objects' appearance, and low diversity of views available for each object.
We remark that these are novel views of object instances never seen during training time, and despite this, our method recovers more accurate images than the baselines, with less noise around high-frequency textures.

\paragraph{Outdoor scenes.}\label{sec:kitti}
{KITTI~\cite{Geiger2012CVPR}} is a widely used standard dataset for outdoor driving scenarios, consisting of complex city scenes.
The dataset contains image sequences as well as the camera poses for each frame.
The transformations in this case are 3D translation and 3D rotation, for a total of 6 parameters.
Following prior work~\cite{chen2019monocular}, in total we have 18,560 and 4,641 images for training and testing, respectively. For each training pair, we randomly select the target view within $\pm5$ frames of the source view.
%
%
For complex scenes like KITTI, reconstructing the target images purely from the latent embedding is challenging.
In order to reconstruct more details, we adapted a warping-based pipeline~\cite{chen2019monocular} that samples pixels from the source input, by inserting our shifted tensor (eq.~\ref{eq:equiv-shift}) between the encoder and decoder.
The corresponding results are shown in table~\ref{tab:quant} and figure~\ref{fig:kitti}, where we highlight some areas where our method recovers details where Chen et al.'s introduces artifacts (the other methods failed to produce meaningful images so we omit them from this figure).

\begin{table}
    \centering
    \begin{minipage}{.49\textwidth}
    \centering
    \caption{Comparison between additive and product tensor-spaces for combining transformations (section \ref{sec:additive-vs-product-expm})}
    \label{tab:eff}
    \begin{adjustbox}{max width=\textwidth}
    \begin{tabular}{@{}lcccccc@{}}
    \toprule
    \multirow{2}{*}{Space} &
      \multicolumn{2}{c}{MNIST~\cite{lecun1998gradient}} &
      \multicolumn{2}{c}{3D Shapes~\cite{3dshapes18}} &
      \multicolumn{2}{c}{Efficiency} \\
     &
      \multicolumn{1}{c}{PSNR} &
      \multicolumn{1}{c}{SSIM} &
      \multicolumn{1}{c}{PSNR} &
      \multicolumn{1}{c}{SSIM} &
      \multicolumn{1}{c}{Mem.} &
      \multicolumn{1}{c}{\#Params.}
      \\ \midrule
    Product  & 23.74 & 0.8772 & 30.77 & 0.9210 & 48GB & 2088M \\ 
    Additive & 21.14 & 0.7942 & 32.09 & 0.9064 & 4.4GB & 205M \\
    \bottomrule
    \end{tabular}
    \end{adjustbox}
    \end{minipage}
    \hfill
    \begin{minipage}{.49\textwidth}
    \centering
    \caption{Results of the compositionality experiment, with unseen concepts (section~\ref{sec:compositionality}).}
    \label{tab:zs}
    \begin{adjustbox}{max width=\textwidth}
    \begin{tabular}{@{}lcccccc@{}}
    \toprule
    Test set &
      \multicolumn{2}{c}{Blue + sphere} &
      \multicolumn{2}{c}{Large + cylinder} &
      \multicolumn{2}{c}{Cube + red wall} \\
    Metric &
      \multicolumn{1}{c}{PSNR} &
      \multicolumn{1}{c}{SSIM} &
      \multicolumn{1}{c}{PSNR} &
      \multicolumn{1}{c}{SSIM} &
      \multicolumn{1}{c}{PSNR} &
      \multicolumn{1}{c}{SSIM}
      \\ \midrule
    Trans. AE~\cite{hinton2011transforming}  & 18.49 & 0.7631 & 17.62 & 0.7025 & 18.34 & 0.7304 \\ 
    Ours & 28.48 & 0.8912 & 24.27 & 0.8270 & 27.81 & 0.8354 \\
    \bottomrule
    \end{tabular}
    \end{adjustbox}
    \end{minipage}
\end{table}

\subsection{Additive vs. product space of transformations}\label{sec:additive-vs-product-expm}
As discussed in section~\ref{s:qae}, we use a rich yet expensive product-space of transformations (eq.~\ref{eq:multi-transf}), but we also consider a more parsimonious and efficient additive space (eq.~\ref{eq:effi}).
We show a comparison of the two schemes in table~\ref{tab:eff} on two datasets.
We can see that the additive space does not cause the performance to drop significantly, and can even improve it slightly (for PSNR on DeepMind 3D Shapes).
On the other hand, the additive space is much more efficient than the product space (in memory consumption and number of parameters) when the transformation space is large (\eg the 6-dimensional space of the 3D Shapes).

\subsection{Relative pose estimation}
The equivariance property of the trained auto-encoders means that they can be used for the downstream task of relative pose estimation (between two images), although they are not directly trained on pose regression.
This can be done by extracting the embedding tensors of two images, and shifting one relative to the other (eq.~\ref{eq:equiv-shift}) to find the highest matching relative pose (measured with a cosine distance between the two embeddings).
To test this idea, we sample image pairs from DeepMind 3D Shapes~\cite{3dshapes18}, where the first image is sampled uniformly, and the second is the first image transformed by an out-of-plane rotation in the range $[-15^\circ, 15^\circ]$.
We now use the above procedure to predict the relative pose between them.
This yields an average error rate over the test set (80,000 images) of just $1.077^\circ$.
Since the angles in the dataset are quantised to $1^\circ$, this is within the expected discretisation error of our model, and represents near-perfect accuracy.
As a comparison, we also test the performance of Trans.AE~\cite{hinton2011transforming} and a direct regression baseline, with the same backbone, which despite our best efforts achieve accuracies of only $4.924^\circ$ and $4.968^\circ$ respectively.

\subsection{Compositionality and out-of-distribution extrapolation}\label{sec:compositionality}
An important property of our proposal is that the transformation factors are forced to be disentangled (eq.~\ref{eq:multi-transf} and eq.~\ref{eq:effi}).
This means that our method may be well equipped to compose transformations in novel ways, extrapolating beyond the training set.
We test this idea by leaving one particular combination of attributes out of the training set (\eg blue spheres are never observed), in DeepMind 3D Shapes.
By training on the remaining samples, we test the network's ability to reconstruct a concept that was never observed before.
The results are in table~\ref{tab:zs}, where our method successfully composes and extrapolates outside the training distribution for 3 pairs of attributes, while a transforming auto-encoder~\cite{hinton2011transforming} does not.

\section{Conclusion}
In this work, we propose a method to train a deep network to achieve equivariance to arbitrary transformations, entirely from data, which goes beyond the typical scenario where an analytical model of the transformations (\eg geometrical) is given.
We demonstrate encouraging results in re-rendering images for given arbitrary transformation parameters, on both synthetic and more complex real datasets.
We also experiment with extrapolating compositions beyond the training set, and a simple relative pose estimation result.
Interesting avenues for future work include investigating ways to increase the visual fidelity, such as with GANs or auto-regressive models, though the simplicity of the current auto-encoder is also attractive.

\subsection*{Acknowledgments}
We are grateful for the support of the EPSRC Programme Grant Visual AI (EP/T028572/1), and the Royal Academy of Engineering (RF\textbackslash 201819\textbackslash 18\textbackslash 163).

\bibliography{egbib}

\begin{thebibliography}{57}
\providecommand{\natexlab}[1]{#1}
\providecommand{\url}[1]{\texttt{#1}}
\expandafter\ifx\csname urlstyle\endcsname\relax
  \providecommand{\doi}[1]{doi: #1}\else
  \providecommand{\doi}{doi: \begingroup \urlstyle{rm}\Url}\fi

\bibitem[Bruna et~al.(2013)Bruna, Szlam, and LeCun]{bruna2013learning}
Joan Bruna, Arthur Szlam, and Yann LeCun.
\newblock Learning stable group invariant representations with convolutional
  networks.
\newblock In \emph{1st International Conference on Learning Representations,
  ICLR}, 2013.

\bibitem[Burgess and Kim(2018)]{3dshapes18}
Chris Burgess and Hyunjik Kim.
\newblock {3D Shapes Dataset}.
\newblock https://github.com/deepmind/3dshapes-dataset/, 2018.

\bibitem[Chaurasia et~al.(2013)Chaurasia, Duchene, Sorkine-Hornung, and
  Drettakis]{chaurasia2013depth}
Gaurav Chaurasia, Sylvain Duchene, Olga Sorkine-Hornung, and George Drettakis.
\newblock Depth synthesis and local warps for plausible image-based navigation.
\newblock \emph{ACM Transactions on Graphics (TOG)}, 32\penalty0 (3):\penalty0
  1--12, 2013.

\bibitem[Chen et~al.(2019)Chen, Song, and Hilliges]{chen2019monocular}
Xu~Chen, Jie Song, and Otmar Hilliges.
\newblock Monocular neural image based rendering with continuous view control.
\newblock In \emph{Proceedings of the IEEE International Conference on Computer
  Vision}, pages 4090--4100, 2019.

\bibitem[Cohen et~al.(2018)Cohen, Geiger, Köhler, and Welling]{s2018spherical}
Taco~S. Cohen, Mario Geiger, Jonas Köhler, and Max Welling.
\newblock Spherical {CNN}s.
\newblock In \emph{International Conference on Learning Representations}, 2018.

\bibitem[Cohen et~al.(2019)Cohen, Geiger, and Weiler]{CohenNEURIPS2019}
Taco~S Cohen, Mario Geiger, and Maurice Weiler.
\newblock A general theory of equivariant cnns on homogeneous spaces.
\newblock In \emph{Advances in Neural Information Processing Systems}, 2019.

\bibitem[Dosovitskiy et~al.(2016)Dosovitskiy, Springenberg, Tatarchenko, and
  Brox]{dosovitskiy2016learning}
Alexey Dosovitskiy, Jost~Tobias Springenberg, Maxim Tatarchenko, and Thomas
  Brox.
\newblock Learning to generate chairs, tables and cars with convolutional
  networks.
\newblock \emph{IEEE transactions on pattern analysis and machine
  intelligence}, 39\penalty0 (4):\penalty0 692--705, 2016.

\bibitem[Dupont et~al.(2020)Dupont, Martin, Colburn, Sankar, Susskind, and
  Shan]{dupont2020equivariant}
Emilien Dupont, Miguel~Bautista Martin, Alex Colburn, Aditya Sankar, Josh
  Susskind, and Qi~Shan.
\newblock Equivariant neural rendering.
\newblock In \emph{International Conference on Machine Learning}, pages
  2761--2770. PMLR, 2020.

\bibitem[Eslami et~al.(2018)Eslami, Rezende, Besse, Viola, Morcos, Garnelo,
  Ruderman, Rusu, Danihelka, Gregor, et~al.]{eslami2018neural}
SM~Ali Eslami, Danilo~Jimenez Rezende, Frederic Besse, Fabio Viola, Ari~S
  Morcos, Marta Garnelo, Avraham Ruderman, Andrei~A Rusu, Ivo Danihelka, Karol
  Gregor, et~al.
\newblock Neural scene representation and rendering.
\newblock \emph{Science}, 360\penalty0 (6394):\penalty0 1204--1210, 2018.

\bibitem[Fitzgibbon et~al.(2005)Fitzgibbon, Wexler, and
  Zisserman]{fitzgibbon2005image}
Andrew Fitzgibbon, Yonatan Wexler, and Andrew Zisserman.
\newblock Image-based rendering using image-based priors.
\newblock \emph{International Journal of Computer Vision}, 63\penalty0
  (2):\penalty0 141--151, 2005.

\bibitem[Flynn et~al.(2016)Flynn, Neulander, Philbin, and
  Snavely]{flynn2016deepstereo}
John Flynn, Ivan Neulander, James Philbin, and Noah Snavely.
\newblock Deepstereo: Learning to predict new views from the world's imagery.
\newblock In \emph{Proceedings of the IEEE conference on computer vision and
  pattern recognition}, pages 5515--5524, 2016.

\bibitem[Flynn et~al.(2019)Flynn, Broxton, Debevec, DuVall, Fyffe, Overbeck,
  Snavely, and Tucker]{flynn2019deepview}
John Flynn, Michael Broxton, Paul Debevec, Matthew DuVall, Graham Fyffe, Ryan
  Overbeck, Noah Snavely, and Richard Tucker.
\newblock Deepview: View synthesis with learned gradient descent.
\newblock In \emph{Proceedings of the IEEE Conference on Computer Vision and
  Pattern Recognition}, pages 2367--2376, 2019.

\bibitem[Geiger et~al.(2012)Geiger, Lenz, and Urtasun]{Geiger2012CVPR}
Andreas Geiger, Philip Lenz, and Raquel Urtasun.
\newblock {Are we ready for autonomous driving? The KITTI Vision Benchmark
  Suite}.
\newblock In \emph{Conference on Computer Vision and Pattern Recognition
  (CVPR)}, 2012.

\bibitem[Goodfellow et~al.(2016)Goodfellow, Bengio, and
  Courville]{goodfellow2016deep}
Ian Goodfellow, Yoshua Bengio, and Aaron Courville.
\newblock \emph{Deep learning}.
\newblock MIT Press, 2016.

\bibitem[Goodfellow et~al.(2014)Goodfellow, Pouget-Abadie, Mirza, Xu,
  Warde-Farley, Ozair, Courville, and Bengio]{goodfellow2014generative}
Ian~J Goodfellow, Jean Pouget-Abadie, Mehdi Mirza, Bing Xu, David Warde-Farley,
  Sherjil Ozair, Aaron Courville, and Yoshua Bengio.
\newblock Generative adversarial networks.
\newblock \emph{arXiv preprint arXiv:1406.2661}, 2014.

\bibitem[Hedman et~al.(2018)Hedman, Philip, Price, Frahm, Drettakis, and
  Brostow]{hedman2018deep}
Peter Hedman, Julien Philip, True Price, Jan-Michael Frahm, George Drettakis,
  and Gabriel Brostow.
\newblock Deep blending for free-viewpoint image-based rendering.
\newblock \emph{ACM Transactions on Graphics (TOG)}, 37\penalty0 (6):\penalty0
  1--15, 2018.

\bibitem[Henriques and Vedaldi(2017)]{henriques2017warp}
J.~F. Henriques and A.~Vedaldi.
\newblock Warped convolutions: Efficient invariance to spatial transformations.
\newblock In \emph{ICML}, 2017.

\bibitem[Henriques et~al.(2014)Henriques, Martins, Caseiro, and
  Batista]{henriques2014pose}
J.~F. Henriques, P.~Martins, R.~Caseiro, and J.~Batista.
\newblock Fast training of pose detectors in the fourier domain.
\newblock In \emph{Advances in Neural Information Processing Systems}, 2014.

\bibitem[Higgins et~al.(2016)Higgins, Matthey, Pal, Burgess, Glorot, Botvinick,
  Mohamed, and Lerchner]{higgins2016beta}
Irina Higgins, Loic Matthey, Arka Pal, Christopher Burgess, Xavier Glorot,
  Matthew Botvinick, Shakir Mohamed, and Alexander Lerchner.
\newblock {Beta-VAE}: Learning basic visual concepts with a constrained
  variational framework.
\newblock 2016.

\bibitem[Higgins et~al.(2018)Higgins, Amos, Pfau, Racaniere, Matthey, Rezende,
  and Lerchner]{higgins2018towards}
Irina Higgins, David Amos, David Pfau, Sebastien Racaniere, Loic Matthey,
  Danilo Rezende, and Alexander Lerchner.
\newblock Towards a definition of disentangled representations.
\newblock \emph{arXiv preprint arXiv:1812.02230}, 2018.

\bibitem[Hinton et~al.(2011)Hinton, Krizhevsky, and
  Wang]{hinton2011transforming}
Geoffrey~E Hinton, Alex Krizhevsky, and Sida~D Wang.
\newblock Transforming auto-encoders.
\newblock In \emph{International conference on artificial neural networks},
  pages 44--51. Springer, 2011.

\bibitem[Hinton et~al.(2018)Hinton, Sabour, and Frosst]{hinton2018matrix}
Geoffrey~E Hinton, Sara Sabour, and Nicholas Frosst.
\newblock Matrix capsules with {EM} routing.
\newblock In \emph{International conference on learning representations}, 2018.

\bibitem[Kanazawa et~al.(2014)Kanazawa, Sharma, and
  Jacobs]{kanazawa2014locally}
Angjoo Kanazawa, Abhishek Sharma, and David Jacobs.
\newblock Locally scale-invariant convolutional neural networks.
\newblock \emph{arXiv preprint arXiv:1412.5104}, 2014.

\bibitem[Kingma and Welling(2014)]{kingma_auto-encoding_2014}
Diederik~P. Kingma and Max Welling.
\newblock Auto-{Encoding} {Variational} {Bayes}.
\newblock In \emph{ICLR}, 2014.

\bibitem[Kondor and Trivedi(2018)]{kondor2018generalization}
Risi Kondor and Shubhendu Trivedi.
\newblock On the generalization of equivariance and convolution in neural
  networks to the action of compact groups.
\newblock In \emph{International Conference on Machine Learning}, pages
  2747--2755, 2018.

\bibitem[Kopf et~al.(2013)Kopf, Langguth, Scharstein, Szeliski, and
  Goesele]{kopf2013image}
Johannes Kopf, Fabian Langguth, Daniel Scharstein, Richard Szeliski, and
  Michael Goesele.
\newblock Image-based rendering in the gradient domain.
\newblock \emph{ACM Transactions on Graphics (TOG)}, 32\penalty0 (6):\penalty0
  1--9, 2013.

\bibitem[Kosiorek et~al.(2019)Kosiorek, Sabour, Teh, and
  Hinton]{kosiorek2019stacked}
Adam Kosiorek, Sara Sabour, Yee~Whye Teh, and Geoffrey~E Hinton.
\newblock Stacked capsule autoencoders.
\newblock In \emph{Advances in Neural Information Processing Systems}, pages
  15512--15522, 2019.

\bibitem[Krizhevsky et~al.(2012)Krizhevsky, Sutskever, and
  Hinton]{krizhevsky2012imagenet}
Alex Krizhevsky, Ilya Sutskever, and Geoffrey~E Hinton.
\newblock Imagenet classification with deep convolutional neural networks.
\newblock In \emph{Advances in neural information processing systems}, pages
  1097--1105, 2012.

\bibitem[Kulkarni et~al.(2015)Kulkarni, Whitney, Kohli, and
  Tenenbaum]{kulkarni2015deep}
Tejas~D Kulkarni, William~F Whitney, Pushmeet Kohli, and Joshua~B Tenenbaum.
\newblock Deep convolutional inverse graphics network.
\newblock In \emph{Proceedings of the 28th International Conference on Neural
  Information Processing Systems-Volume 2}, pages 2539--2547, 2015.

\bibitem[Ladicky et~al.(2014)Ladicky, Shi, and Pollefeys]{ladicky2014pulling}
Lubor Ladicky, Jianbo Shi, and Marc Pollefeys.
\newblock Pulling things out of perspective.
\newblock In \emph{Proceedings of the IEEE conference on computer vision and
  pattern recognition}, pages 89--96, 2014.

\bibitem[Lai et~al.(2011)Lai, Bo, Ren, and Fox]{lai2011large}
Kevin Lai, Liefeng Bo, Xiaofeng Ren, and Dieter Fox.
\newblock A large-scale hierarchical multi-view {RGB-D} object dataset.
\newblock In \emph{2011 IEEE international conference on robotics and
  automation}, pages 1817--1824. IEEE, 2011.

\bibitem[LeCun et~al.(1998)LeCun, Bottou, Bengio, and
  Haffner]{lecun1998gradient}
Yann LeCun, L{\'e}on Bottou, Yoshua Bengio, and Patrick Haffner.
\newblock Gradient-based learning applied to document recognition.
\newblock \emph{Proceedings of the IEEE}, 86\penalty0 (11):\penalty0
  2278--2324, 1998.

\bibitem[LeCun et~al.(2004)LeCun, Huang, and Bottou]{lecun2004learning}
Yann LeCun, Fu~Jie Huang, and Leon Bottou.
\newblock Learning methods for generic object recognition with invariance to
  pose and lighting.
\newblock In \emph{Proceedings of the 2004 IEEE Computer Society Conference on
  Computer Vision and Pattern Recognition, 2004. CVPR 2004.}, volume~2, pages
  II--104. IEEE, 2004.

\bibitem[Lenc and Vedaldi(2015)]{lenc2015understanding}
Karel Lenc and Andrea Vedaldi.
\newblock Understanding image representations by measuring their equivariance
  and equivalence.
\newblock In \emph{Proceedings of the IEEE conference on computer vision and
  pattern recognition}, pages 991--999, 2015.

\bibitem[Matusik et~al.(2002)Matusik, Pfister, Ngan, Beardsley, Ziegler, and
  McMillan]{matusik2002image}
Wojciech Matusik, Hanspeter Pfister, Addy Ngan, Paul Beardsley, Remo Ziegler,
  and Leonard McMillan.
\newblock Image-based {3D} photography using opacity hulls.
\newblock \emph{ACM Transactions on Graphics (TOG)}, 21\penalty0 (3):\penalty0
  427--437, 2002.

\bibitem[Memisevic and Hinton(2010)]{memisevic2010learning}
Roland Memisevic and Geoffrey~E Hinton.
\newblock Learning to represent spatial transformations with factored
  higher-order {Boltzmann} machines.
\newblock \emph{Neural computation}, 22\penalty0 (6):\penalty0 1473--1492,
  2010.

\bibitem[Miao and Rao(2007)]{miao_learning_2007}
Xu~Miao and Rajesh~PN Rao.
\newblock Learning the lie groups of visual invariance.
\newblock \emph{Neural computation}, 19\penalty0 (10):\penalty0 2665--2693,
  2007.

\bibitem[Mildenhall et~al.(2019)Mildenhall, Srinivasan, Ortiz-Cayon, Kalantari,
  Ramamoorthi, Ng, and Kar]{mildenhall2019local}
Ben Mildenhall, Pratul~P Srinivasan, Rodrigo Ortiz-Cayon, Nima~Khademi
  Kalantari, Ravi Ramamoorthi, Ren Ng, and Abhishek Kar.
\newblock Local light field fusion: Practical view synthesis with prescriptive
  sampling guidelines.
\newblock \emph{ACM Transactions on Graphics (TOG)}, 38\penalty0 (4):\penalty0
  1--14, 2019.

\bibitem[Mildenhall et~al.(2020)Mildenhall, Srinivasan, Tancik, Barron,
  Ramamoorthi, and Ng]{mildenhall2020nerf}
Ben Mildenhall, Pratul~P Srinivasan, Matthew Tancik, Jonathan~T Barron, Ravi
  Ramamoorthi, and Ren Ng.
\newblock Nerf: Representing scenes as neural radiance fields for view
  synthesis.
\newblock In \emph{European Conference on Computer Vision}, pages 405--421.
  Springer, 2020.

\bibitem[Park et~al.(2017)Park, Yang, Yumer, Ceylan, and
  Berg]{park2017transformation}
Eunbyung Park, Jimei Yang, Ersin Yumer, Duygu Ceylan, and Alexander~C Berg.
\newblock Transformation-grounded image generation network for novel {3D} view
  synthesis.
\newblock In \emph{Proceedings of the ieee conference on computer vision and
  pattern recognition}, pages 3500--3509, 2017.

\bibitem[Penner and Zhang(2017)]{penner2017soft}
Eric Penner and Li~Zhang.
\newblock Soft 3d reconstruction for view synthesis.
\newblock \emph{ACM Transactions on Graphics (TOG)}, 36\penalty0 (6):\penalty0
  1--11, 2017.

\bibitem[Qi et~al.(2020)Qi, Zhang, Lin, and Wang]{qi2020learning}
Guo-Jun Qi, Liheng Zhang, Feng Lin, and Xiao Wang.
\newblock Learning generalized transformation equivariant representations via
  autoencoding transformations.
\newblock \emph{IEEE Transactions on Pattern Analysis and Machine
  Intelligence}, 2020.

\bibitem[Reddy and Chatterji(1996)]{reddy1996fft}
B.~Srinivasa Reddy and Biswanath~N. Chatterji.
\newblock An {FFT}-based technique for translation, rotation, and
  scale-invariant image registration.
\newblock \emph{IEEE Transactions on Image Processing}, 5\penalty0
  (8):\penalty0 1266--1271, 1996.

\bibitem[Rematas et~al.(2016)Rematas, Nguyen, Ritschel, Fritz, and
  Tuytelaars]{rematas2016novel}
Konstantinos Rematas, Chuong~H Nguyen, Tobias Ritschel, Mario Fritz, and Tinne
  Tuytelaars.
\newblock Novel views of objects from a single image.
\newblock \emph{IEEE transactions on pattern analysis and machine
  intelligence}, 39\penalty0 (8):\penalty0 1576--1590, 2016.

\bibitem[Schmidt and Roth(2012)]{schmidt_learning_2012}
Uwe Schmidt and Stefan Roth.
\newblock Learning rotation-aware features: {From} invariant priors to
  equivariant descriptors.
\newblock In \emph{Computer {Vision} and {Pattern} {Recognition} ({CVPR}), 2012
  {IEEE} {Conference} on}, pages 2050--2057, 2012.

\bibitem[Seitz et~al.(2006)Seitz, Curless, Diebel, Scharstein, and
  Szeliski]{seitz2006comparison}
Steven~M Seitz, Brian Curless, James Diebel, Daniel Scharstein, and Richard
  Szeliski.
\newblock A comparison and evaluation of multi-view stereo reconstruction
  algorithms.
\newblock In \emph{2006 IEEE computer society conference on computer vision and
  pattern recognition (CVPR'06)}, volume~1, pages 519--528. IEEE, 2006.

\bibitem[Simard et~al.(2000)Simard, Le~Cun, Denker, and
  Victorri]{simard2000transformation}
Patrice~Y Simard, Yann~A Le~Cun, John~S Denker, and Bernard Victorri.
\newblock Transformation invariance in pattern recognition: Tangent distance
  and propagation.
\newblock \emph{International Journal of Imaging Systems and Technology},
  11\penalty0 (3):\penalty0 181--197, 2000.

\bibitem[Sitzmann et~al.(2019)Sitzmann, Zollh{\"o}fer, and
  Wetzstein]{sitzmann2019srns}
Vincent Sitzmann, Michael Zollh{\"o}fer, and Gordon Wetzstein.
\newblock Scene representation networks: Continuous {3D}-structure-aware neural
  scene representations.
\newblock In \emph{Advances in Neural Information Processing Systems}, 2019.

\bibitem[Srinivasan et~al.(2019)Srinivasan, Tucker, Barron, Ramamoorthi, Ng,
  and Snavely]{srinivasan2019pushing}
Pratul~P Srinivasan, Richard Tucker, Jonathan~T Barron, Ravi Ramamoorthi, Ren
  Ng, and Noah Snavely.
\newblock Pushing the boundaries of view extrapolation with multiplane images.
\newblock In \emph{Proceedings of the IEEE Conference on Computer Vision and
  Pattern Recognition}, pages 175--184, 2019.

\bibitem[Sun et~al.(2018)Sun, Huh, Liao, Zhang, and Lim]{sun2018multi}
Shao-Hua Sun, Minyoung Huh, Yuan-Hong Liao, Ning Zhang, and Joseph~J Lim.
\newblock Multi-view to novel view: Synthesizing novel views with self-learned
  confidence.
\newblock In \emph{Proceedings of the European Conference on Computer Vision
  (ECCV)}, pages 155--171, 2018.

\bibitem[Tatarchenko et~al.(2016)Tatarchenko, Dosovitskiy, and
  Brox]{tatarchenko2016multi}
Maxim Tatarchenko, Alexey Dosovitskiy, and Thomas Brox.
\newblock Multi-view 3d models from single images with a convolutional network.
\newblock In \emph{European Conference on Computer Vision}, pages 322--337.
  Springer, 2016.

\bibitem[Tenenbaum and Freeman(2000)]{tenenbaum2000separating}
Joshua~B Tenenbaum and William~T Freeman.
\newblock Separating style and content with bilinear models.
\newblock \emph{Neural computation}, 12\penalty0 (6):\penalty0 1247--1283,
  2000.

\bibitem[Van~Oord et~al.(2016)Van~Oord, Kalchbrenner, and
  Kavukcuoglu]{van2016pixel}
Aaron Van~Oord, Nal Kalchbrenner, and Koray Kavukcuoglu.
\newblock Pixel recurrent neural networks.
\newblock In \emph{International Conference on Machine Learning}, pages
  1747--1756. PMLR, 2016.

\bibitem[Wu et~al.(2020)Wu, Rupprecht, and Vedaldi]{Wu_2020_CVPR}
Shangzhe Wu, Christian Rupprecht, and Andrea Vedaldi.
\newblock Unsupervised learning of probably symmetric deformable {3D} objects
  from images in the wild.
\newblock In \emph{Proceedings of the IEEE/CVF Conference on Computer Vision
  and Pattern Recognition (CVPR)}, June 2020.

\bibitem[Xiao et~al.(2019)Xiao, Liu, and Lee]{xiao2019identity}
Fanyi Xiao, Haotian Liu, and Yong~Jae Lee.
\newblock Identity from here, pose from there: Self-supervised disentanglement
  and generation of objects using unlabeled videos.
\newblock In \emph{Proceedings of the IEEE/CVF International Conference on
  Computer Vision}, pages 7013--7022, 2019.

\bibitem[Yang et~al.(2015)Yang, Reed, Yang, and Lee]{yang2015weakly}
Jimei Yang, Scott~E Reed, Ming-Hsuan Yang, and Honglak Lee.
\newblock Weakly-supervised disentangling with recurrent transformations for 3d
  view synthesis.
\newblock In \emph{Advances in Neural Information Processing Systems}, pages
  1099--1107, 2015.

\bibitem[Zhou et~al.(2016)Zhou, Tulsiani, Sun, Malik, and Efros]{zhou2016view}
Tinghui Zhou, Shubham Tulsiani, Weilun Sun, Jitendra Malik, and Alexei~A Efros.
\newblock View synthesis by appearance flow.
\newblock In \emph{European conference on computer vision}, pages 286--301.
  Springer, 2016.

\end{thebibliography}

\end{document}